\documentclass[10pt,twocolumn,letterpaper]{article}

\usepackage{cvpr}              

\usepackage[utf8]{inputenc} 
\usepackage[T1]{fontenc}    
\usepackage{hyperref}       
\usepackage{url}            
\usepackage{booktabs}       
\usepackage{amsfonts}       
\usepackage{amsmath}
\usepackage{nicefrac}       
\usepackage{microtype}      
\usepackage{adjustbox}
\usepackage{multirow}
\usepackage{xcolor}
\usepackage{graphics,psfrag}
\usepackage{diagbox} 
\usepackage{bbding}
\usepackage{amssymb}
\usepackage{times}
\usepackage{epsfig}
\usepackage{algorithmic}
\usepackage{algorithm2e}
\usepackage{float}
\usepackage{rotating}
\usepackage{makecell}
\usepackage{pifont}
\definecolor{ForestGreen}{rgb}{0.13, 0.55, 0.13}
\definecolor{Maroon}{rgb}{0.69, 0.19, 0.0}
\newcommand{\cmark}{\ding{51}}%

\newcolumntype{C}[1]{>{\centering\arraybackslash}m{#1}}
\newcolumntype{R}[1]{>{\raggedleft\arraybackslash}m{#1}}
\newcolumntype{P}[1]{>{\raggedright\arraybackslash}p{#1}}

\newcolumntype{M}[1]{>{\centering\arraybackslash}m{#1}}
\def\eg{\emph{e.g.}}

\usepackage[capitalize]{cleveref}
\crefname{section}{Sec.}{Secs.}
\Crefname{section}{Section}{Sections}
\Crefname{table}{Table}{Tables}
\crefname{table}{Tab.}{Tabs.}

\def\cvprPaperID{3967} 
\def\confName{CVPR}
\def\confYear{2022}

\begin{document}

\title{Look Back and Forth: Video Super-Resolution with Explicit Temporal Difference Modeling}
\author{%
	Takashi Isobe$^{1}$ \qquad Xu Jia$^2$\thanks{Corresponding author} \qquad Xin Tao$^1$ \qquad Changlin Li$^1$ \qquad Ruihuang Li$^3$ \\ Yongjie Shi$^4$ \qquad Jing Mu$^1$ \qquad Huchuan Lu$^{2,5}$ \qquad Yu-Wing Tai$^{1}$\footnotemark[1]\\
	{$^1$Kuaishou Technology} \qquad
	{$^2$Dalian University of Technology}\\
	{$^3$Hong Kong Polytechnic University} \qquad
    {$^4$Peking University} \qquad
    {$^5$Peng Cheng Laboratory}
    \\
	{\texttt{\small{\{isobetakashi, taoxin, lichanglin, mujing03, daiyurong\}}@kuaishou.com}\hspace{0.5cm}} \\
	{\texttt{\small{\{xjia, lhchuan\}}@dlut.edu.cn } \hspace{0.5cm}} 
	{\texttt{\small{csrhli}@comp.polyu.edu.hk}\hspace{0.5cm}} 
	{\texttt{\small{shiyongjie}@pku.edu.cn}\hspace{0.5cm}}
}
\maketitle

\begin{abstract}
Temporal modeling is crucial for video super-resolution. Most of the video super-resolution methods adopt the optical flow or deformable convolution for explicitly motion compensation. However, such temporal modeling techniques increase the model complexity and might fail in case of occlusion or complex motion, resulting in serious distortion and artifacts. In this paper, we propose to explore the role of explicit temporal difference modeling in both LR and HR space. Instead of directly feeding consecutive frames into a VSR model, we propose to compute the temporal difference between frames and divide those pixels into two subsets according to the level of difference. They are separately processed with two branches of different receptive fields in order to better extract complementary information. To further enhance the super-resolution result, not only spatial residual features are extracted, but the difference between consecutive frames in high-frequency domain is also computed. It allows the model to exploit intermediate SR results in both future and past to refine the current SR output. The difference at different time steps could be cached such that information from further distance in time could be propagated to the current frame for refinement. Experiments on several video super-resolution benchmark datasets demonstrate the effectiveness of the proposed method and its favorable performance against state-of-the-art methods.


\end{abstract}
\section{Introduction}
\label{intro}

Super-resolution (SR) is an important vision task that aims at recovering high-resolution (HR) images from low-resolution (LR) observations. Single image super-resolution (SISR)~\cite{dong2014learning,kim2016accurate,kim2016deeply,Lai-cvpr17-LapSRN,haris2018deep,zhang2018image,mu2020graph,chan2021glean,wei2021unsupervised} methods reply much on image prior learned from large datasets or self-similarity within images to synthesize high frequency contents, while video super-resolution (VSR) methods~\cite{liu2017robust,wang2019deformable,isobe2020revisiting,yi2021omniscient,jing2021turning,li2021comisr,cao2021video,lee2021dynavsr} are expected to extract valuable complementary details from neighbouring frames, which could provide more information to alleviate the ill-posed problem.
Both of these tasks have achieved remarkable progress thanks to the development of deep learning techniques.  
As more and more videos are recorded, VSR has become a key component in many applications such as video remastering, live streaming and surveillance. 
\begin{figure}[t]
	\centering
	\includegraphics[width=1\columnwidth]{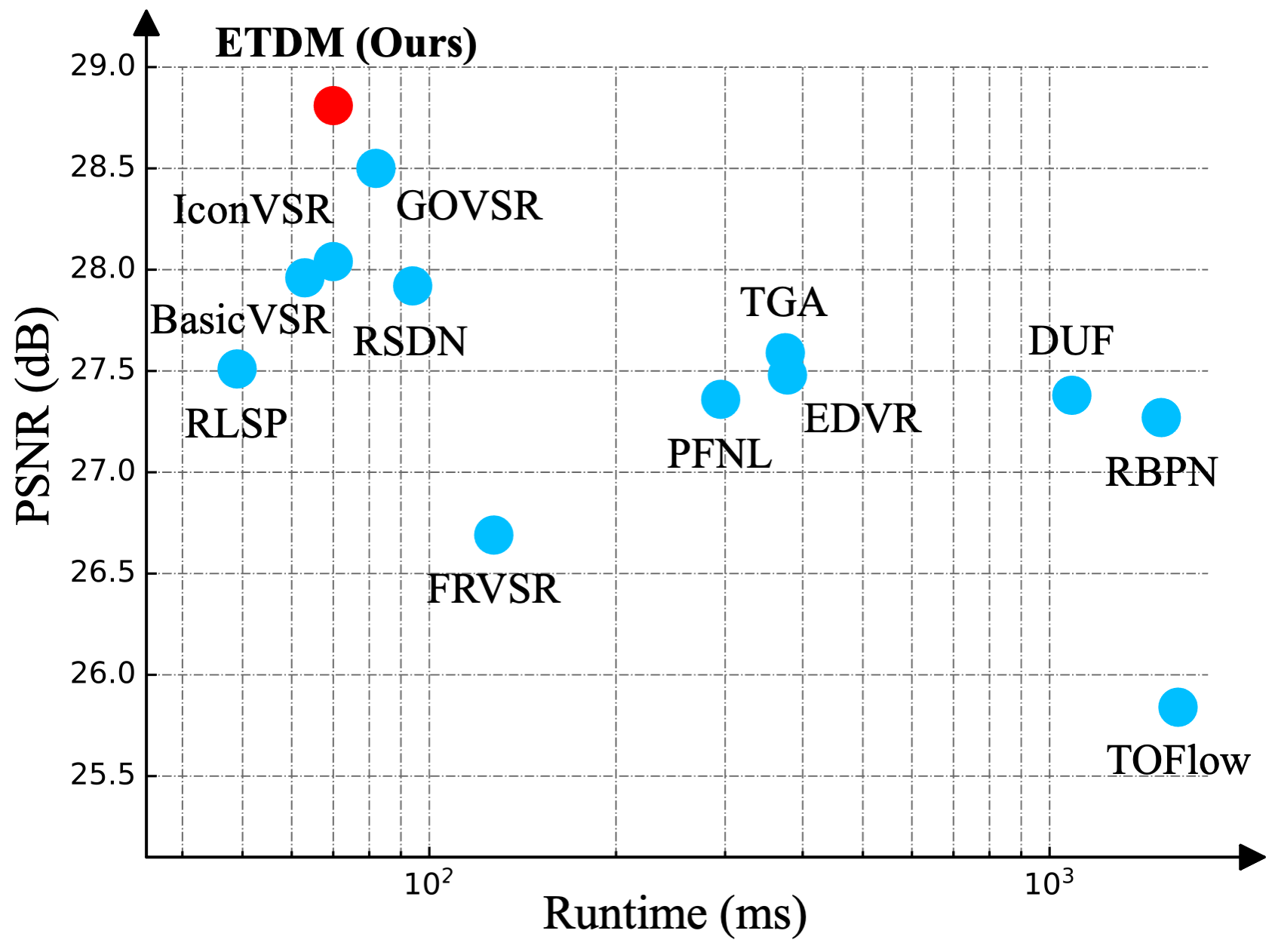}
	\vspace{-2mm}
	\caption{VSR performance comparison on Vid4~\cite{liu2013bayesian} in terms of PSNR (dB) and runtime (ms). Our proposed ETDM outperforms the previous methods with high efficiency.}
	\label{fig:temporal_structure}
	\vspace{-3mm}
\end{figure}

To effectively explore abundant spatial-temporal information within frames, some methods~\cite{li2019fast,xue2019video,wang2019deformable,wang2019edvr,chan2020understanding,tian2020tdan,ying2020deformable} attempt to model temporal information across frames through explicit or implicit motion compensation. However, explicit motion compensation~\cite{xue2019video,wang2019edvr} would increase the model complexity, and inevitable errors in motion estimation may cause distortion and degrade super-resolution results. The methods with implicit motion compensation, \textit{e.g.,} 3D convolutional layer~\cite{jo2018deep,liu2021large}, bet all on model capacity, ignoring the valuable temporal priors. Another kind of method explores temporal information following a recurrent fashion in either uni-direction or bi-direction. They can accumulate rich history information in the hidden state, either only from  past~\cite{sajjadi2018frame,Fuoli-arxiv19-rlsp,isobe2020video-eccv,isobe2020revisiting,liu2017robust}, or from both future and past to extract beneficial complementary information for detail recovery~\cite{chan2020basicvsr, chan2021basicvsr++,yi2021omniscient}. However, they suffer from either unbalanced historical accumulation at each frame or large memory caching.

Although many techniques have been proposed to extract complementary information, the differences among frames and SR results of different time steps have not been explicitly explored yet. Very recently, the idea of explicit temporal difference modeling was explored and was successfully applied to video-related tasks either to improve its performance or efficiency. In~\cite{wang2016temporal,wang2021tdn}, authors proposed to exploit the RGB difference between frames as an efficient alternative to optical flow to model motion. The temporal difference network is able to effectively capture both short-term and long-term information, which is essential in the action recognition task. 

In this work, we explore the role of explicit temporal difference modeling in both LR and HR space. The VSR is conducted in a uni-directional recurrent way for efficiency and avoiding large memory caching of bi-direction.
Instead of directly feeding consecutive frames into a VSR model, we propose to compute the temporal difference between the reference frame and the neighboring frame. The neighboring frames are divided into two subsets according to the level of difference, with smaller ones as low variance regions and larger ones as high variance regions. They are separately fed together with the reference frame into two branches with different receptive fields. The outputs of these two branches are combined and fed to the \textit{Spatial-Residual Head} to reconstruct the initial SR results. Additionally, the \textit{Future-Residual Head} and \textit{Past-Residual Head} are respectively used to model temporal difference in HR space based on the initial spatial residual features at future and past time steps. In this way, the result at the current step would be further enhanced in HR space by allowing the model to look back and forth at intermediate estimation in both future and past. In addition, the temporal difference between spatial
residual features at different time steps would be cached. Therefore, information from further time steps could all be propagated to the current time step for comprehensive refinement. Compared to the bi-directional based methods, the proposed method could enjoy both the efficiency of uni-directional network and the power of bi-directional information propagation, but with a flexible cache.
The proposed method achieves favorable performance against state-of-the-art methods on several benchmark datasets. Several ablation studies are conducted to examine the effectiveness of its components.

Our main contributions are as follows: (1) A new framework to explicitly explore the temporal difference in both LR and HR for the VSR task; (2) A novel back-and-forth refinement strategy to boost performance; (3) Favorable performance against state-of-the-arts on several VSR benchmarks.

\section{Related Work}\label{related}
\noindent\textbf{Multi-frame super-resolution.} 
Some methods explore the temporal information in an explicit way. Xue~\textit{et al.}~\cite{xue2019video} proposed a new sub-pixel motion compensation layer to calculate the optical flow and perform up-sampling simultaneously. TDAN~\cite{tian2020tdan} and EDVR~\cite{wang2019edvr} adopt deformable convolutions to work in the feature level motion alignment. However, methods with explicit motion compensation suffer from huge computational costs and inevitable errors in motion estimation may be prone to generate artifacts. To avoid these issues,~\cite{jo2018deep,isobe2020video,yi2019progressive} conduct VSR with implicit motion compensation. Jo~\textit{et al.}~\cite{jo2018deep} proposed to use 3D CNN for estimating a dynamic upsampling filter. In~\cite{yi2019progressive},~Yi~\textit{et al.} proposed a non-local extraction module to model spatial-temporal correlations between the neighboring frame and reference frame. MuCAN~\cite{li2020mucan} selects and fuses top-K most similar patches across frames based on patch-based matching strategy. These methods with well-designed modules achieve promising results but inevitably increase the runtime and model complexity. Different from these works, we propose to explicitly compute the temporal difference across frames in LR space for better handling the complementary information from low-variance and high-variance regions. The proposed method is highly effective in combining multi-frame temporal information for details recovery. 

\noindent\textbf{Recurrent networks for video super-resolution.} 
Another kind of method attempts to exploit the long-range temporal information in a recurrent manner. FRVSR~\cite{sajjadi2018frame} propagates the last built high-resolution frame recurrently. RSLP~\cite{Fuoli-arxiv19-rlsp} introduces high dimensional latent states to propagate temporal information implicitly. RSDN~\cite{isobe2020video-eccv} adopts dual-branch to address different difficulties of structure and detail components in VSR. These methods only propagate the previously estimated results to the current time step for restoration, which leads to imbalanced information for different frames. Generally, the first frame suffers from severe distortion. \cite{huang2015bidirectional,chan2020basicvsr,chan2021basicvsr++} propagate both estimated results from past and future by maintaining two hidden states in a bi-directional way. However, these methods have to take the whole sequence as input, which is memory-consuming and not appropriate for real-time tasks like live broadcasts. LOVSR~\cite{yi2021omniscient} provides a compromised strategy based on the uni-directional recurrent network, which first uses one network to generate the hidden state at the next time step and propagate it back to another network at the current time step for reconstruction. Besides, the SR results are not explicitly used for refinement.

Different from~\cite{yi2021omniscient}, our proposed ETDM explicitly models the temporal difference between adjacent time steps in HR space, such that the intermediate SR results at the past and future could be propagated to the current time step with a single network. Furthermore, accumulating the temporal difference between adjacent time steps could propagate the SR results at arbitrary time steps to the current step for comprehensive refinement. Overall, the key contribution of this paper is to effectively leverage the temporal difference in LR and HR space to restore the fine details.

\begin{figure*}[t]
	\centering
	\includegraphics[width=0.99\textwidth]{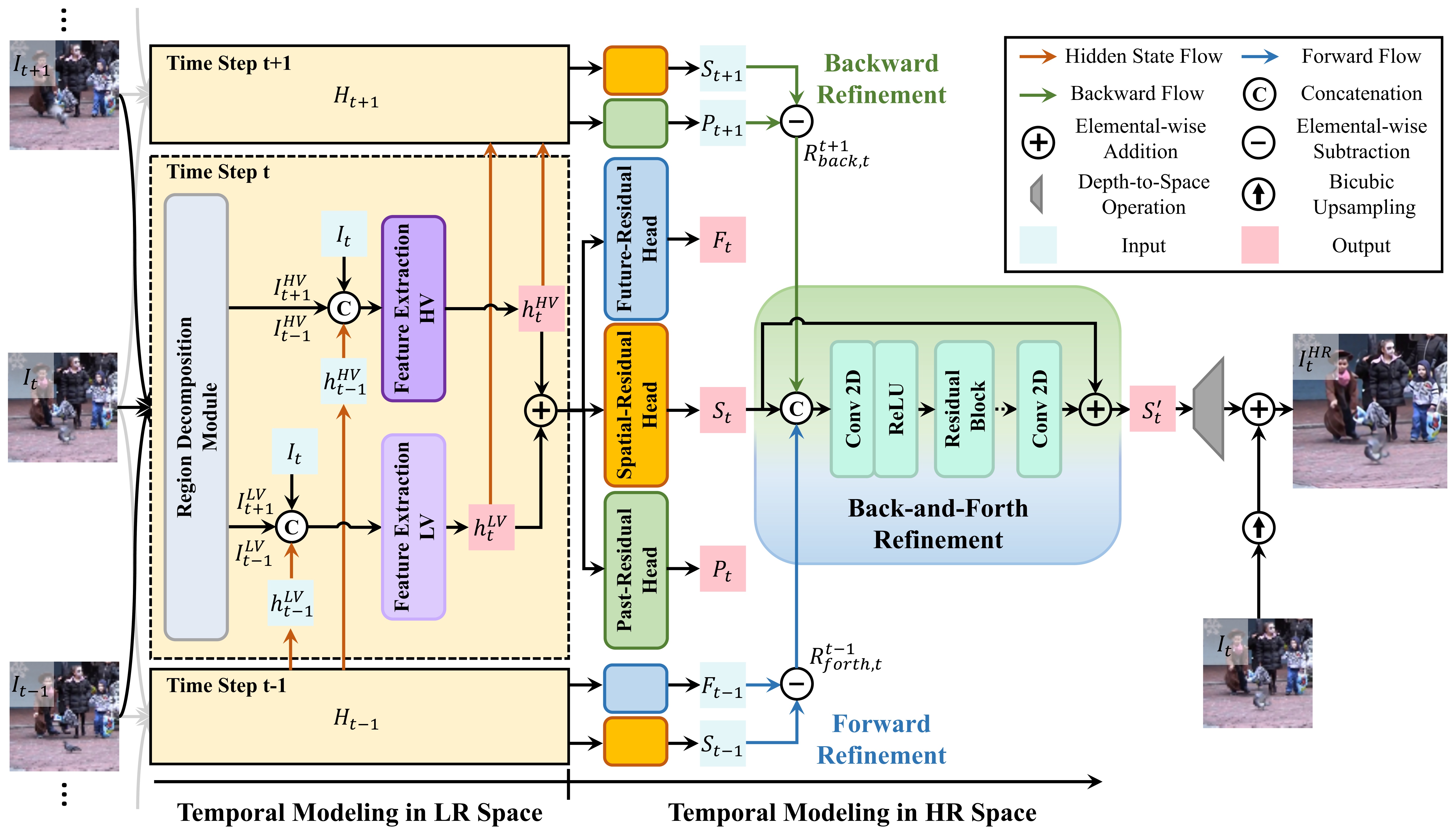}
	\caption{Pipeline of the proposed framework of Explicit Temporal Difference Modeling (ETDM), which conducts VSR in a uni-directional manner. By explicitly modeling the temporal difference in LR and HR space, the proposed method is able to make full use of the complementary information across frames and SR results from past and future time steps. }
	\label{fig:ours}
\end{figure*}

\section{Method}
\label{method}
\subsection{Overview}

In this work, VSR is conducted in a uni-directional recurrent way. For each time step, the network takes the neighboring frames ${I_{t-1}, I_{t}, I_{t+1}}$ and previously estimated SR results as input. The key of the proposed method is to explicitly model the temporal difference in both LR and HR space. 
Formally, we denote $I_t$ as the reference frame, and the temporal difference is defined by the difference between $I_{t}$ and the neighboring frame $I_{t\pm1}$.
The overview of the proposed pipeline is shown in Figure~\ref{fig:ours}. Our proposed method consists of temporal difference modeling in LR space and HR space. In LR space, the proposed Region Decomposition Module computes the difference between the reference frame and neighboring frame. Further, it decomposes the neighboring frames into low-variance (LV) and high-variance (HV) regions according to the level of difference. They are then processed separately by two CNN branches with different receptive fields for better extracting complementary information.

We also encourage the model to predict the temporal difference between SR outputs at adjacent time steps in HR space, which allows the super-resolution at the current step to benefit from initial SR results from both past and future time steps.
In addition, it is natural to extend forward and backward propagation from one time step only to arbitrary time order by caching all temporal differences between two specified time steps.

 

\begin{figure}[t]
	\centering
	\includegraphics[width=0.99\linewidth]{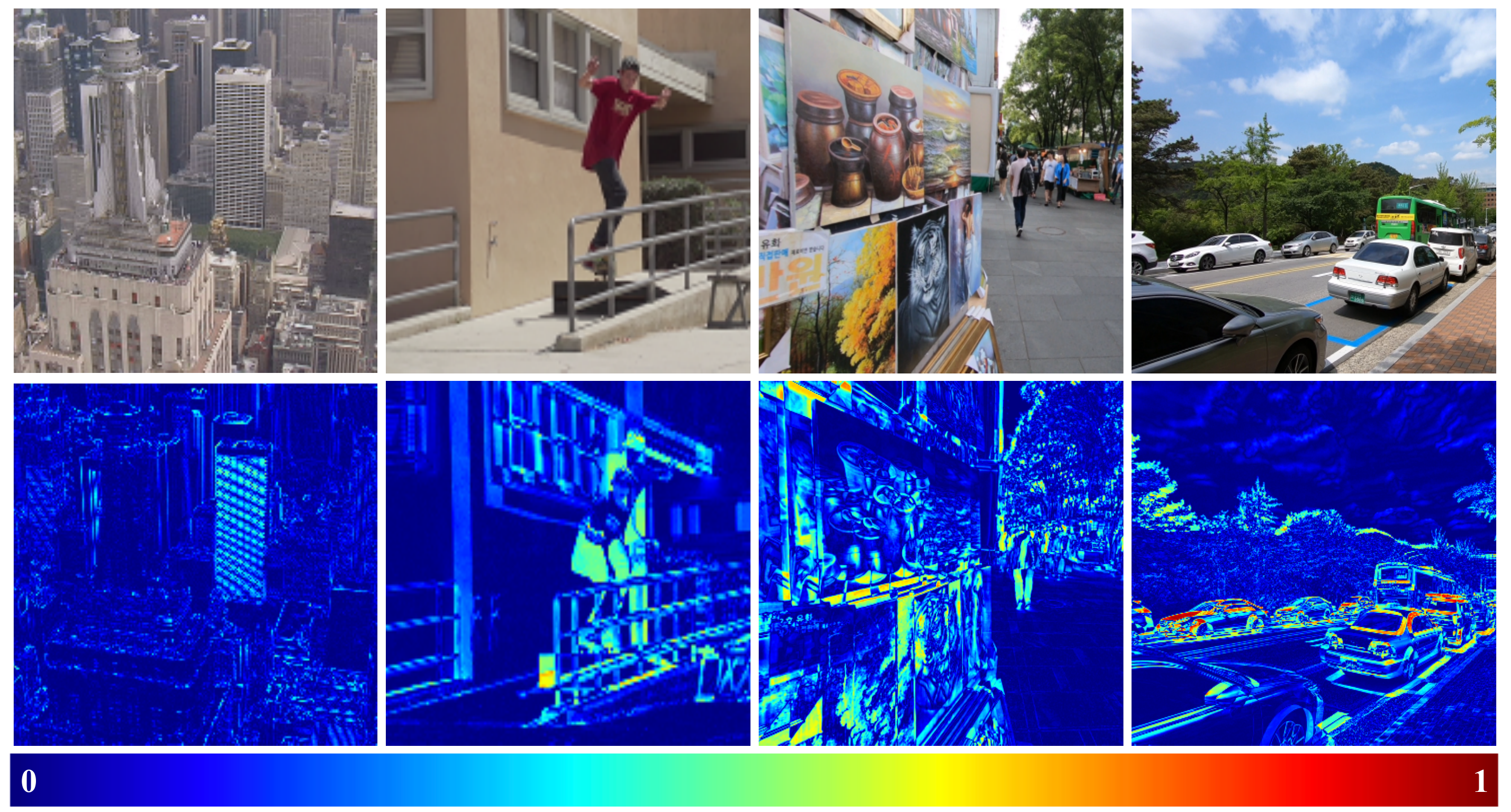}
	\caption{Illustration of the pixel-wise difference maps between two consecutive frames. Color denotes the level of difference.
	}
	\label{fig:diff}
	\vspace{2mm}
\end{figure}

\subsection{Explicit Temporal Difference Modeling}
\noindent\textbf{Temporal difference.}
The goal of video super-resolution is to take advantage of complementary information from the neighboring frames to reconstruct richer details for the reference frame.
    Figure~\ref{fig:diff} shows different levels of variance with different regions, which motivates us to divide the regions of neighboring frames into low-variance (LV) and high-variance (HV) ones according to the level of temporal difference. The overall appearance is slightly changed in LV regions. Thus the main difference across frames lies in the fine details. As for the HV regions, the overall appearance varies much between frames and may provide coarse-scale complementary information from different perspectives.

%
Here, a median filter with $3\times3$ kernel size is applied to the binarized temporal difference maps and the result is further processed by a set of morphological operations to obtain difference masks for LV regions $M^{LV}$.
Simultaneously, the difference masks for HV regions can be obtained by $M^{HV}=1-M^{LV}$. The LV regions and HV regions of neighboring frame $I_{t-1}$ and $I_{t+1}$ can be then obtained respectively by following Eq.~\ref{eq:mask_t1} and Eq.~\ref{eq:mask_t2}.

\begin{align}
     I^{LV}_{t-1} = M^{LV}_{t-1} \odot I_{t-1},\quad I^{HV}_{t-1} = M^{HV}_{t-1} \odot I_{t-1}, \label{eq:mask_t1}\\ 
     I^{LV}_{t+1} = M^{LV}_{t+1} \odot I_{t+1},\quad I^{HV}_{t+1} = M^{HV}_{t+1} \odot I_{t+1},
\label{eq:mask_t2}
\end{align}
where $\odot$ denotes element-wise multiplication. 
Due to the smoothness with a natural image, the LV regions more likely correspond to the regions with small motion between frames, while the HV regions may correspond to the regions with large motion. Therefore, they should be processed by separate models with different receptive fields.
%
%

\vspace{0.05in}
\noindent\textbf{Temporal modeling in LR space.} 
Here we only take the branch for LV regions at time step $t$ as an example to explain its model design. 
The input of the LV region branch is the masked frames corresponding to consecutive time steps are $\{I^{LV}_{t-1}, I_t, I^{LV}_{t+1}\}$ , and the hidden state $h^{LV}_{t-1}$ at previous time step. 
They are concatenated and then further processed by a convolutional layer and several residual blocks. In this way, this recurrent unit $H_t$ manages to aggregate complementary information from regions with small variance and motion over time. The branch for HV regions is designed in a similar way but all convolutional layers are equipped with the dilation rate two to handle the possible large motion with a larger receptive field.
The output of the LV and HV branch is denoted as $h^{LV}_{t}$ and $h^{HV}_{t}$, respectively.

\vspace{0.05in}
\noindent\textbf{Temporal modeling in HR space.}
In addition, we also calculate the temporal difference in HR space for further refinement. The temporal difference in HR space builds a bridge between the adjacent time steps, such that information is able to propagate to the current time step for refinement.  The output of each branch $h^{LV}_{t-1}$ and $h^{HV}_{t-1}$ are combined and fed to three residual heads, those are, \textit{Spatial-Residual Head}, \textit{Past-Residual Head} and \textit{Future-Residual Head}. 
The Spatial-Residual Head is designed to calculate spatial residual between bicubic-upsampled reference frame and the high-resolution ground-truth, denoted as $S_t$. 
The Future-Residual Head computes the spatial residual between the corresponding high-resolution temporal difference $(I^{HR}_{t}-I^{HR}_{t+1})$ and the bicubic-upsampled temporal difference $(I^{\uparrow}_{t} - I^{\uparrow}_{t+1})$, which is also equivalent to the temporal difference between spatial residual at different time steps, denoted as $F_t$ in Eq.~\ref{eq:fut_res}. 
\begin{equation}
\begin{aligned}
F_t & =  (I^{HR}_{t}-I^{HR}_{t+1}) -(I^{\uparrow}_{t} - I^{\uparrow}_{t+1})
\\ & = (I^{HR}_{t} - I^{\uparrow}_{t}) - (I^{HR}_{t+1}-I^{\uparrow}_{t+1}).
\label{eq:fut_res}
\end{aligned}
\end{equation}
Similarly, the Past-Residual Head computes the spatial residual of the temporal difference  $(I^{HR}_{t}-I^{HR}_{t-1}) -(I^{\uparrow}_{t} - I^{\uparrow}_{t-1})$ denoted as $P_t$. 
By using the temporal difference between adjacent time steps in HR space, the initial SR estimation at past and future time steps could be propagated to the current time step to refine its SR result. 

\begin{figure}[t]
	\centering
  \begin{center}
  \includegraphics[width=0.5\textwidth]{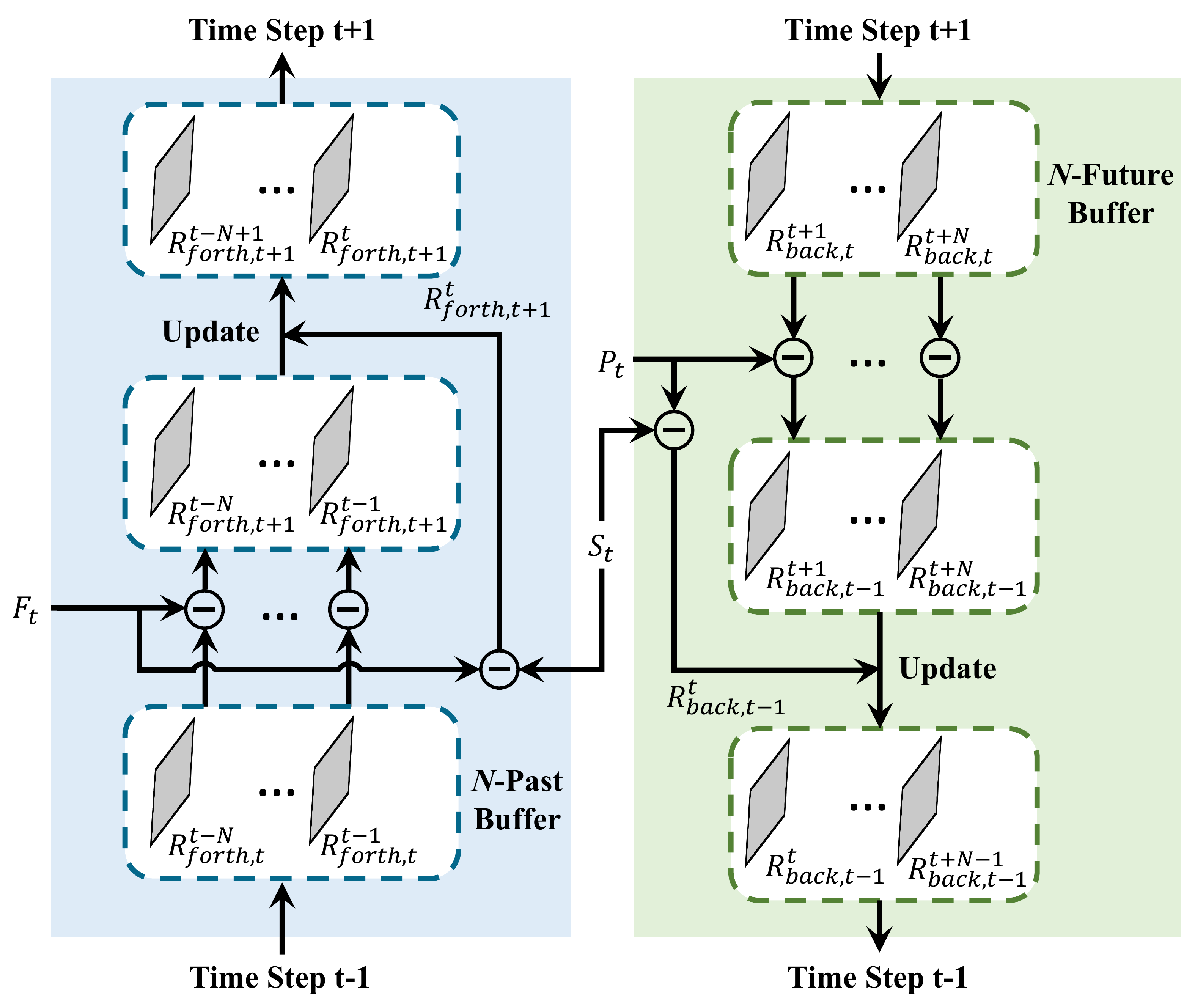}	
  \end{center}  
  \vspace{-5mm}
  \caption{\small Illustration of update of N-Past and N-Future buffer at time step t.}
  \label{fig:refinement}
\end{figure}

\subsection{Back-and-Forth Refinement}
\label{sec:refinement}
In this section, we will detail how the temporal difference in HR space and estimation at other time steps could help refine the SR result at the current time step.
%

Promising VSR results of bi-directional based methods~\cite{chan2020basicvsr, chan2021basicvsr++, yi2021omniscient} could be credited to its bi-directional propagation, which allows the model to aggregate information from the whole sequence. However, it has to cache all intermediate hidden states, limiting its application in many scenarios. In this paper, the proposed method also allows to propagate bidirectional information to enhance the current frame, but it is conducted with only a uni-directional recurrent network without the need for a large cache.
Specifically, with the predicted temporal difference in HR space, the neighboring time steps $S_{t-1}$ and $S_{t+1}$ can be respectively propagated to the current step as follows:
\begin{equation}
    R^{t-1}_{forth,t} = S_{t-1}-F_{t-1}, \quad
    R^{t+1}_{back,t}= S_{t+1}-P_{t+1},
    \label{eq:ref_def_p1}
\end{equation}
where $R^{t-1}_{forth,t}$ and $R^{t+1}_{back,t}$ respectively denote the propagated spatial residual from past and future to the current time step $t$.
For $R^{m}_{forth,n}$ ($m\textless n$), the superscript represents the time step of used information in the past and the subscript represents the target time it forwards to. For $R^{k}_{back,n}$ ($k \textgreater n$), the superscript the represents the 
time step of used information in the future and the subscript represents the target time it backwards propagates to. 
To further refine the current SR output with information propagated from other time steps, we concatenate $R^{t-1}_{forth,t}$, $R^{t+1}_{back,t}$ and $S_t$ as input to a convolutional layer followed by several residual blocks to obtain the refined spatial residual $S^{'}_{t}$. The final super-resolved image is generated by adding the pixel-shuffled $S^{'}_t$ to the bicubic-upsampled reference frame.



\vspace{0.05in}
\noindent\textbf{Extension to arbitrary temporal order refinement.}
It is natural to extend forward and backward propagation from one time step only to arbitrary time order $l$ by accumulating the temporal difference across several time steps. For example, the forward propagation from time step $(t-l)$ to $t$ can be formulated as below:
\begin{equation}
R^{t-l}_{forth,t}= S_{t-l} - (\sum\limits^{\substack{l}}_{i=1} F_{t-i}).
\label{eq:ref_def_p2}
\end{equation}
Similarly, we can also propagate the spatial residual from future time step $(t+l)$ to $t$,
\begin{equation}
R^{t+l}_{back,t} = S_{t+l} - (\sum\limits_{\substack{i=1}}^{l} P_{t+i}).
    \label{eq:ref_def_f1}
\end{equation}


In order to make full use of the information propagated from different time steps to the current one, we maintain N-Past Buffer and N-Future Buffer of size $N$ to cache the desired intermediate results \textit{i.e.,} $\{R^{t-l}_{forth,t}, l=1,\cdots, N\}$ and $\{R^{t+l}_{back,t}, l=1,\cdots, N\}$ for forward and backward refinement, respectively.

The spatial residual $S_t$ would be further refined using all elements in N-Past and N-Future Buffers in a similar way as that with one time step cache explained before.

\vspace{0.05in}
\noindent\textbf{Buffer update.}
Once the final SR result at time step $t$ is obtained, the recurrent model would conduct the same super-resolution operation for frame $I_{t+1}$. In this case, the model would require not only updated hidden state but also updated buffer that caches all intermediate spatial residuals from different time steps.
The buffer update follows the First-in-First-out principle, that is, the oldest intermediate result $R^{t-N}_{forth,t+1}$ and $R^{t+N}_{back,t-1}$ are respectively removed from N-Past Buffer and N-Future Buffer. 
While a new intermediate result $R^{t}_{forth,t+1}$ and $R^{t}_{back,t-1}$ are respectively added to the two buffers.
As for the remained elements in the buffer, their update proceeds as follows,
\begin{equation}
    R^{m}_{forth,t+1} = R^{m}_{forth,t} - F_{t}, \quad
    R^{k}_{back,t-1} = R^{k}_{back,t} - P_{t}.
    \label{eq:buffer_update}
\end{equation}
%
\subsection{Losses}
Similar to most VSR works~\cite{wang2019deformable,chan2020basicvsr,yi2019progressive}, where only reconstruction loss is used to supervise the VSR model's training. However, supervision here comes from both spatial reconstruction and temporal reconstruction.
%
For each time step, we compute the difference between both initially estimated and refined spatial residual and the ground-truth as spatial reconstruction losses.
\begin{equation}
    \mathcal{L}^\mathcal{S}_t = \sqrt{\Vert {S}_{t}^{GT} - {S}_t  \Vert^2 + \varepsilon^2},~ 
    \mathcal{L}^\mathcal{S^{'}}_t =  \sqrt{\Vert {S}_{t}^{GT} - S_t^{'} \Vert^2 + \varepsilon^2},
\end{equation}
where ${S}_{t}^{GT}$ is the ground-truth spatial residual and $\varepsilon$ is empirically set to $1\times 10^{-3}$.
Since the spatial refinement is computed based on the spatial residual estimation from multiple time steps away in both past and future, stricter supervision is indirectly imposed on the parameters of the model.
In addition, the temporal residuals are also supervised by the corresponding ground-truth,
\begin{equation}
    \mathcal{L}^\mathcal{F}_t =  \sqrt{\Vert {F}_{t}^{GT}-{F}_t   \Vert^2 + \varepsilon^2},~
    \mathcal{L}^\mathcal{P}_t = \sqrt{\Vert {P}_{t}^{GT}-{P}_t   \Vert^2 + \varepsilon^2}. 
\end{equation}
The total loss $\mathcal{L}$ can be computed as below,
\begin{equation}
\mathcal{L} = \dfrac{1}{N} \sum_{t=1}^{N}(\mathcal{L}^\mathcal{F}_t + \mathcal{L}^\mathcal{P}_t + \mathcal{L}^\mathcal{S}_t + \mathcal{L}_t^{\mathcal{S}^{'}}).
\label{eq:4}
\end{equation}

\section{Experiments}\label{sec:exp}
\label{experiment}

\subsection{Implementation Details}
\noindent\textbf{Datasets.}
Some previous works train their VSR models on the private dataset, which is not suitable for a fair comparison. In this work, we adopt the popular used Vimeo-90K~\cite{xue2019video} as our training set, which consists of about $90K$ 7-frame video clips with various motion types. During training, we randomly sample regions with the patches of size $256\times 256$ from the HR video sequences as the target. Similar to~\cite{Fuoli-arxiv19-rlsp,yi2019progressive,isobe2020video-eccv,yi2021omniscient,chan2021basicvsr++}, the corresponding low-resolution patches of $64\times64$ are produced by applying Gaussian blur with $\sigma=1.6$ to the target patches followed by $\times 4$ downsampling. We conduct our evaluation on four widely used benchmark datasets, including Vid4~\cite{liu2013bayesian}, SPMCS~\cite{tao2017detail} and UDM10~\cite{yi2019progressive}. The SR results are evaluated in terms of PSNR and SSIM on the Y channel of YCbCr space. The PSNR and SSIM are measured excluding the first and the last one frames. We also train and evaluate the proposed method on REDS with the same down-sampling setting used in the NTIRE challenge.

\begin{table}[t]
\footnotesize
\setlength\tabcolsep{3.5pt}
\caption{Ablation studies of the proposed ETDM with the buffer size of $N=3$. "R" is the region decomposition module. "F" and "B" mean the forward and backward refinement, respectively.}\label{tab:ablation}
\begin{center}
\scalebox{1}{	\begin{tabular}{C{1.1cm}|C{0.4cm} C{0.4cm} C{0.4cm} | C{1.0cm} | C{1.5cm} | C{1.5cm} }
		\toprule
		Model \# &R &F &B &\#Param. &Vid4~\cite{liu2013bayesian} &UDM10~\cite{yi2019progressive} \\
		\midrule
		1 & & &   &8.0M	&27.89 &39.46  \\
		\midrule
		2 &\cmark &  &	 & 8.0M	 &28.04 &39.68 \\
		3 &\cmark &\cmark  &	 &8.2M	&28.29  &39.88 \\ 
		4 &\cmark &\cmark  &\cmark   &8.4M &28.81  &40.11\\
		\bottomrule
	\end{tabular}
	}
	\label{table:ablation_CCL}
\end{center}
\end{table}
\begin{figure}[t]
	\centering
	\includegraphics[width=1\columnwidth]{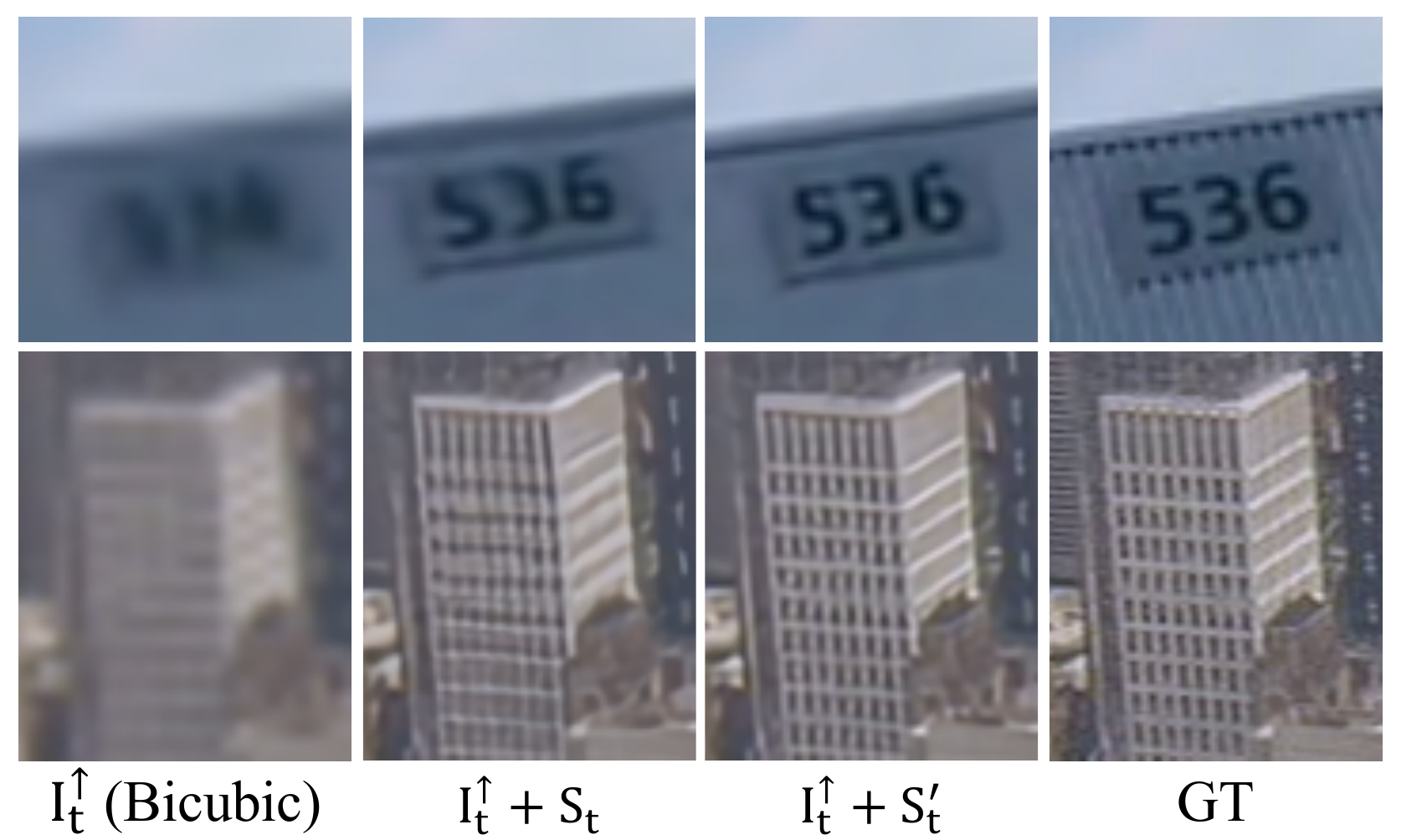}
\caption{\small Visualization of the intermediate and refined SR results. The proposed back-and-forth refinement produces sharper edges and finer detailed textures.}
  \label{fig:backforth}
 	\vspace{2mm}
\end{figure}
\vspace{-2mm}

\vspace{0.05in}
\noindent\textbf{Implementation details.}
The proposed ETDM adopts $2$ residual blocks for LV and HV branches where each convolutional layer has $96$ channels. $16$ residual blocks are used for further feature extraction. To alleviate the large motion in previously estimated hidden state, ETDM adopts the  optical flow~\cite{ranjan2017optical} to perform the spatial alignment on features, similar to~\cite{chan2020basicvsr,fuoli2022fast,lin2021fdan}. 
We adopt $16$ residual blocks with $64$ channels for back-and-forth refinement. To effectively utilize all given frames, we pad each sequence with the reflecting of the first frame and last frame at the beginning and end of the sequence, respectively. The elements in N-Past and N-Future Buffer are respectively initialized with zeros for conducting VSR at the first frame. 
The models are supervised with Charbonnier penalty loss function~\cite{charbonnier1994two} and optimized with Adam optimizer~\cite{kingma2014adam} by setting $\beta_1=0.9$ and $\beta_2=0.999$. Each mini-batch consists of $16$ samples.
The initial learning rate is set to $1 \times 10^{-4}$ for VSR model and $2.5 \times 10^{-5}$ for optical flow estimator. We adopt Cosine Annealing scheme~\cite{loshchilov2016sgdr}. The total number of epochs is 90. The training data is augmented by standard flipping and rotating and an additional temporal reverse operation. All experiments are conducted on a server with Python 3.6.4, PyTorch 1.1 and V100 GPUs.

\subsection{Ablation Study}

\noindent\textbf{Ablation on the components of ETDM.}
In this section, we examine the effectiveness of each component of the proposed ETDM framework on Vid4~\cite{liu2013bayesian} and UDM10~\cite{yi2019progressive} test set, as shown in Table~\ref{table:ablation_CCL}. For a fair comparison, these models have a similar number of parameters. A baseline (Model 1) is designed by taking the original consecutive frames as the two branches' input without performing region decomposition. Model 1 achieves 27.89 dB and 39.46 dB on Vid4 and UDM10, respectively. By dividing the neighboring frames into LV and HV regions, \textit{i.e.,}~Model 2, surpasses Model 1 by +0.15dB and +0.22dB on Vid4 and UDM10, respectively. The improvement would be credited to the model that better utilizes the complementary information from the LV and HV regions at the corresponding branch. By explicitly modeling the temporal difference in HR space, Model 3 leverage the propagated SR results for refinement and gains +0.25dB and +0.20dB over Model 2. By further propagating the SR result from future to the current time step, Model 4 achieves notable improvement by +0.52dB and +0.23dB over Model 3 on Vid4 and UDM10. We also visualize the intermediate and the refined SR results in~Figure~\ref{fig:backforth}. The proposed back-and-forth refinement can produce finer details and stronger edges.


\begin{table}[t]
	\footnotesize
	\setlength\tabcolsep{2.3pt}
	
	\caption{Comparison of the proposed temporal difference modeling in HR space with optical flow (OF) for the back-and-forth refinement. The Runtime is calculated on an HR image size of $1280\times 720$. }
	\vspace{-1mm}
	\begin{center}
		\scalebox{0.9}{
		\begin{tabular}{C{1.7cm}| C{1.2cm} | C{1.2cm} C{1.2cm}|C{1.2cm} C{1.2cm} }
			\toprule
			  &{\multirow{2}{*}{$N=0$}}  &\multicolumn{2}{c|}{$N=1$} &\multicolumn{2}{c}{$N=3$}\\
			 \cmidrule{3-6}
			 & &OF~\cite{ranjan2017optical}  & ETDM & OF~\cite{ranjan2017optical}  & ETDM  \\
			\midrule
			Runtime (ms)  &62 &96  &67 &163  &70
			\\
			Vid4~\cite{liu2013bayesian}   &28.04 &28.29    &28.18  &28.54 &28.81   \\
			UDM10~\cite{yi2019progressive} &39.68 &39.76   &39.77   &40.01  &40.11  \\
			\bottomrule
		\end{tabular}
		}
    	\vspace{-2mm}
		\label{table:flow}
	\end{center}
\end{table}

\vspace{0.05in}
\noindent\textbf{Ablation on the temporal modeling in HR space.}
We adopt the SPyNet~\cite{ranjan2017optical} as an alternative temporal modeling technique in HR space for back and forth refinement, where the past and future estimated SR are aligned to the current time step based on the estimated optical flow in HR space. 
In this experiment, the buffer size of $N=1$ and $N=3$ are used to examine the difference between two kinds of temporal modeling methods.
As shown in Table~\ref{table:flow}, both refinement with optical flow and temporal difference could bring performance gain over baseline. However, our method is more efficient than optical flow in temporal modeling. By looking into further away time steps \textit{i.e.,} $N=3$, our method outperforms the baseline model by 0.77dB and 0.43dB on Vid4 and UDM10, respectively, with a little time increase. The method with optical flow becomes 2 times slower than our method in this case. Besides, its performance is also a little inferior to the method with explicit temporal difference modeling difference in HR space. One possible reason could be its inaccurate motion estimation and alignment which results in distortion and errors, hence deteriorating the final super-resolution performance. 

\vspace{0.05in}

%
\begin{figure}[t]
	\centering
  \begin{center}
  \includegraphics[width=0.5\textwidth]{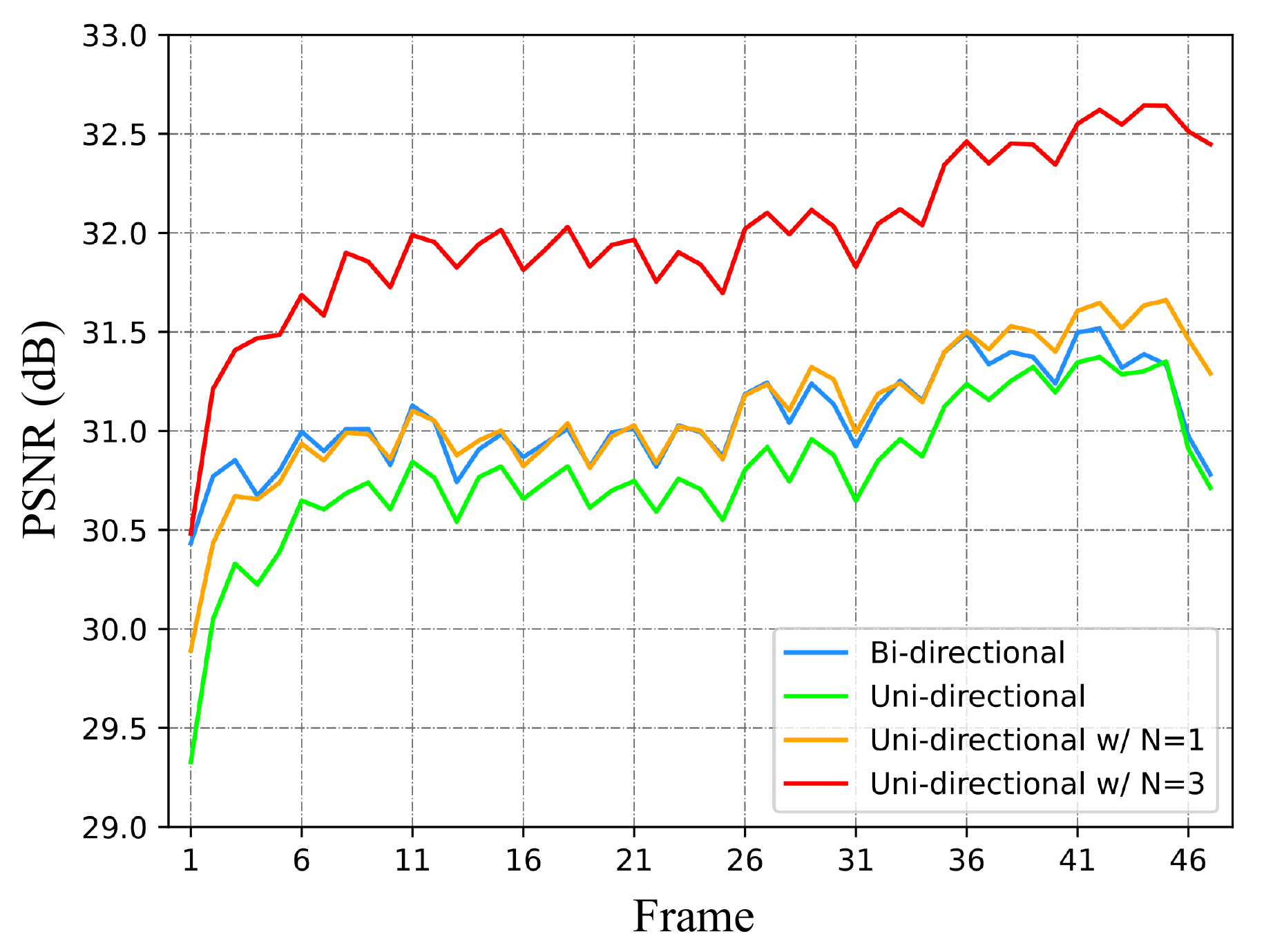}	
  \end{center}  
  \vspace{-3.5mm}
  \caption{\small 
  Results of methods with different information propagation strategies over time. }
	\vspace{-1.5mm}

  \label{fig:propagate}
\end{figure}
To examine the effectiveness of the proposed method, we compare the proposed back and forth propagation with other kinds of uni-directional propagation~\cite{isobe2020revisiting} and bi-directional propagation~\cite{chan2020basicvsr} methods. For a fair comparison, we remove the LV and HV decomposition step of the proposed model and try to keep the number of its parameters the same as the other two methods. As shown in Figure~\ref{fig:propagate}, the bi-directional method ({\color{blue}blue} line) significantly outperforms the uni-directional method ({\color{green}green} line) at first few time steps, but achieves comparable performance at the last two frames. With one step back and forth refinement, \textit{i.e.,} $N=1$, our ablated method performs a little worse than the bi-direction at the first few time steps and outperforms it after the middle of the sequence.
Note that the bi-directional method would have to cache the hidden states of the whole video frames, which will suffer from memory issues when a video contains a lot of frames.

\vspace{0.05in}
\noindent\textbf{Discussion.}
The uni-directional recurrent model stores historical information in the hidden state to complement the details for the current time step. The bi-directional recurrent model adopts two hidden states which not only aggregate the information from past time steps but also the information from future time steps, hence performing better than the uni-directional recurrent model. 
However, both of them only resort to the hidden states that implicitly compress all past or future information into one representation, where some specific complementary information that is important to the current frame might be missing.
Our method explicitly models the temporal difference between the neighboring time step, which not only leverage the specific complementary information from the neighboring time steps for refinement but can also enhance the current estimation using the initial results from more time steps based on the cached temporal difference.


\begin{table*}[t]
	\centering
	\caption{Quantitative comparison (PSNR (dB) and SSIM) on \textbf{Vid4~\cite{liu2013bayesian}}, \textbf{SPMCS~\cite{tao2017detail}}, \textbf{UDM10~\cite{yi2019progressive}} and \textbf{REDS4~\cite{nah2019ntire}} for $4\times$ VSR. {\color{red}Red} text indicates the best and {\color{blue} blue} text indicates the second best performance. The Runtime is calculated on an HR image size of $1280\times 720$. `$\dagger$' means the values are either taken from paper or calculated using provided models. 'uni' and 'bi' represents uni-directional and bi-directional, respectively.} 
	\scalebox{0.98}{	
		\begin{tabular}{l||c|c|c||c|c|c|c}
			\hline
			\textbf{Method}  &\#Frame &Params (M) &Runtime (ms) &\textbf{Vid4~\cite{liu2013bayesian}} &\textbf{SPMCS~\cite{tao2017detail}}  &\textbf{UDM10~\cite{yi2019progressive}} &\textbf{REDS4~\cite{nah2019ntire}}
			\\
			\hline \hline  	
			Bicubic & 1 &N/A &N/A &21.80/0.5426 &23.29/0.6385 &28.47/0.8253 & 26.14/0.7292
			\\	
			TOFlow~\cite{xue2019video} & 7  &1.4 &1610 &25.85/0.7659 &27.86/0.8237	  &36.26/0.9438 &27.93/0.7997
			\\
			DUF~\cite{jo2018deep} & 7   &5.8  &1086  &27.38/0.8329 &29.63/0.8719	 &38.48/0.9605 &28.66/0.8262
			\\
			RBPN~\cite{haris2019recurrent} & 7 &12.2 &1513 &27.27/0.8285  &29.54/0.8704	 &38.56/0.9605 &30.15/0.8593
			\\
			EDVR~\cite{wang2019edvr} & 7  &20.6 &378 &27.85/0.8503 &-/-  &39.89/0.9686  &31.09/0.8800
			\\	
			PFNL~\cite{yi2019progressive} &7  &3.0  &295 &27.36/0.8385  &30.02/0.8804  &38.88/0.9636 &29.63/0.8502
			\\
			TGA~\cite{isobe2020video} &7  &5.8  &375 &27.59/0.8419   &30.31/0.8857 &39.19/0.9645 &-/-
			\\
			FDAN$^\dagger$~\cite{lin2021fdan} &7   &9.0 &- &27.88/0.8508 &-/- &39.91/0.9686 &-/-
            \\ 	\hline \hline 
			FRVSR~\cite{sajjadi2018frame} &uni  &5.1 &137 &26.69/0.8103 &28.16/0.8421	 &37.09/0.9522 &-/-
			\\
			RLSP~\cite{Fuoli-arxiv19-rlsp} &uni &4.2 &49  &27.51/0.8396 &29.64/0.8791  &38.50/0.9614 &30.47/0.8685
			\\
			RSDN~\cite{isobe2020video-eccv} &uni  &6.2 &94 &27.92/0.8505 &30.18/0.8811  &39.35/0.9653 &-/-
			\\
			RRN~\cite{isobe2020revisiting} &uni   &3.4 &45 &27.69/0.8488 &29.84/0.8827 &38.96/0.9644 &-/- 
            \\
            DAP$^\dagger$~\cite{fuoli2022fast} &uni   &- &38 &-/- &-/- &39.50/0.9664  &32.39/0.9069
            \\
			GOVSR~\cite{yi2021omniscient} &bi &7.1 &81 &{\color{blue}28.47}/{\color{blue}0.8722} &{\color{blue}30.34}/{\color{blue}0.8981}	 &{\color{blue}40.08}/{\color{blue}0.9703} &{\color{blue}31.91}/{\color{blue}0.8990}
			\\
			BasicVSR$^\dagger$~\cite{chan2020basicvsr} &bi &6.3  &63 &27.96/0.8553 &-/-  &39.96/0.9694 &31.42/0.8909
			\\
			IconVSR$^\dagger$~\cite{chan2020basicvsr} &bi &8.7 &70 &28.04/0.8570 &-/-  &40.03/0.9694 &31.67/0.8948
			\\
		    \textbf{ETDM} &uni&8.4 &70 &{\color{red}28.81}/{\color{red}0.8725} &{\color{red}30.48}/{\color{red}0.8972}  &{\color{red}40.11}/{\color{red}0.9707} &{\color{red}32.15}/{\color{red}0.9024}
	    	\\
			\hline
		\end{tabular}
	}
 	\vspace{2mm}
	\label{vid4_table}
\end{table*}

\begin{figure*}[t]
	\centering
	\includegraphics[width=0.99\linewidth]{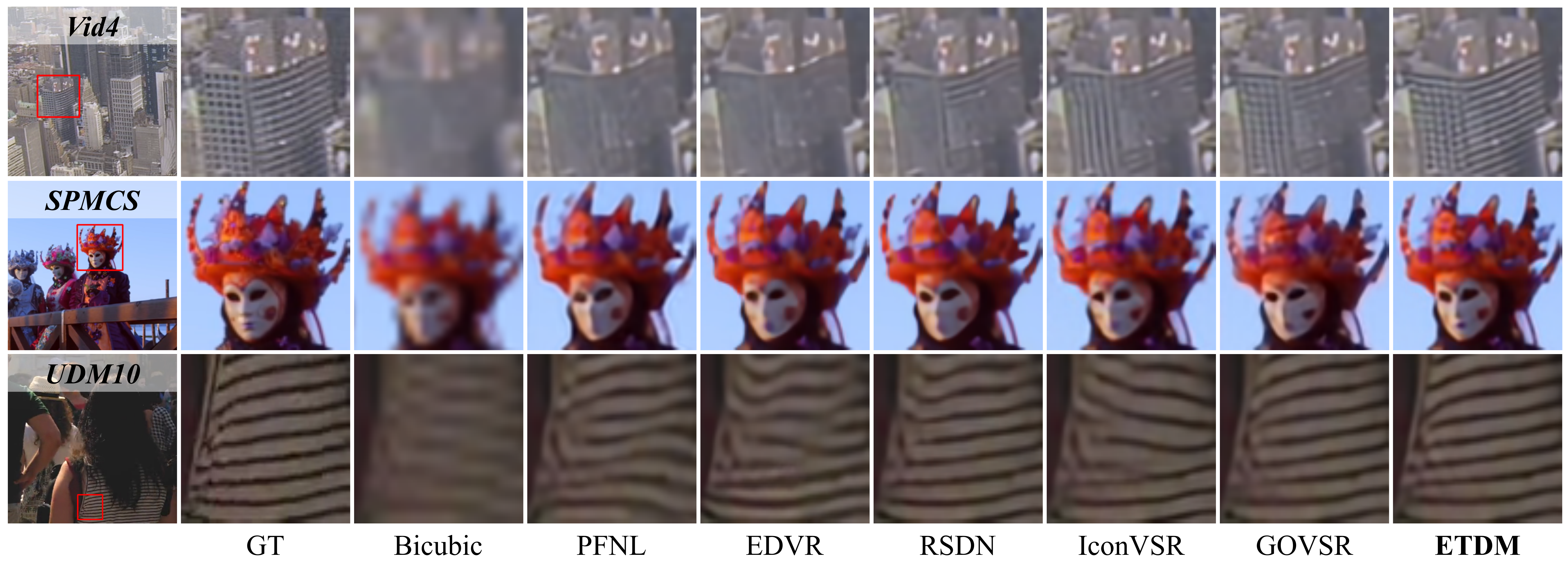}
	\caption{Qualitative comparison on \textbf{Vid4~\cite{liu2013bayesian}}, \textbf{SPMCS~\cite{tao2017detail}} and \textbf{UDM10~\cite{yi2019progressive}} test set for $4\times$ VSR. Zoom in for better visualization.}
	\label{fig:qualitative}
 	\vspace{2mm}
\end{figure*}

\subsection{Comparison with State-of-the-Arts}
In this section, we compare our methods with several state-of-the-art VSR approaches, including TOFlow~\cite{xue2019video}, DUF~\cite{jo2018deep}, RBPN~\cite{haris2019recurrent}, EDVR~\cite{wang2019edvr}, PFNL~\cite{yi2019progressive},
TGA~\cite{isobe2020video}, FDAN~\cite{lin2021fdan},
FRVSR~\cite{sajjadi2018frame}, RLSP~\cite{Fuoli-arxiv19-rlsp}, RSDN~\cite{isobe2020video-eccv}, RRN~\cite{isobe2020revisiting}, DAP~\cite{fuoli2022fast}, GOVSR~\cite{yi2021omniscient}, BasicVSR~\cite{chan2020basicvsr} and IconVSR~\cite{chan2020basicvsr}. The first seven methods conduct VSR within a local window, \textit{e.g.,} 7 frames.
Among these methods, TOFlow, RBPN and EDVR estimate the motion between the reference frame and neighboring frames explicitly. DUF, PFNL, TGA and FDAN conduct VSR with implicit motion compensation. FRVSR, RLSP, RSDN, RRN and DAP super-resolve each frame in a uni-directional recurrent way. GOVSR, IconVSR and BasicVSR conduct VSR in a bi-directional manner. 

There is some difference among these methods on the original training setting.
Different down-sampling kernels are used in the original work of TOFlow, RBPN and EDVR, while DUF, PFNL, FRVSR, RLSP and GOVSR models are trained on different datasets. Therefore, for a fair comparison, we try out best to rebuild these models under the same training setting based on the publicly available code. In addition, we also re-train most of these VSR methods on REDS~\cite{nah2019ntire} following the same down-sampling setting used in the NTIRE challenge.
The quantitative results of the state-of-the-art methods are shown in Table~\ref{vid4_table}. 
ETDM achieves a good balance between speed and reconstruction quality on these datasets. The proposed ETDM with the buffer size of $N=3$ achieves 14 fps in processing a 320$\times$180 video sequence for $4\times $ super-resolution. It is about 4 times faster than the multi-frame super-resolution method EDVR. Moreover, our proposed ETDM with uni-directional hidden state propagation also outperforms the bi-directional-based methods, \textit{i.e.,}~IconVSR and GOVSR which take the whole video sequence as input and have to memorize all intermediate SR results produced by forward propagation. 

The qualitative comparison with other state-of-the-art methods is shown in Figure~\ref{fig:qualitative}. Our method produces higher-quality HR images on three datasets, including finer details and sharper edges. Other methods are either prone to generate some artifacts (\eg, wrong stripes on clothes) or can not recover missing details (\eg, small windows of the building).

\section{Conclusion}
\label{conclusion}
\vspace{-2mm}
In this work, we propose a novel uni-directional recurrent network by explicit temporal difference modeling in both LR and HR space. To temporal modeling in LR space, we propose to compute temporal difference between input frames and divide it into two subsets according to the level of difference. They are separately fed together with the reference frame into two branches with different receptive fields for better extracting the complementary information. To further refine the SR result, we also calculate temporal difference in HR space, which build a bridge between the SR results at adjacent time steps, such that the intermediate SR result at past and future time step could propagate to the current time step for refinement. Extensive experiments demonstrate that the proposed method outperforms existing state of the arts, including recent bi-directional recurrent based methods. 

\textbf{Acknowledgement}
The research was partially supported by the Natural Science Foundation of China, No. 62106036, 61725202, U1903215, and the Fundamental Research Funds for the Central University of China, DUT No. 82232026.

{\small
\bibliographystyle{ieee_fullname}
\bibliography{res}

\begin{thebibliography}{10}\itemsep=-1pt

\bibitem{cao2021video}
Jiezhang Cao, Yawei Li, Kai Zhang, and Luc Van~Gool.
\newblock Video super-resolution transformer.
\newblock {\em CoRR}, abs/2106.06847, 2021.

\bibitem{chan2021glean}
Kelvin~CK Chan, Xintao Wang, Xiangyu Xu, Jinwei Gu, and Chen~Change Loy.
\newblock Glean: Generative latent bank for large-factor image
  super-resolution.
\newblock In {\em CVPR}, 2021.

\bibitem{chan2020basicvsr}
Kelvin~CK Chan, Xintao Wang, Ke Yu, Chao Dong, and Chen~Change Loy.
\newblock Basicvsr: The search for essential components in video
  super-resolution and beyond.
\newblock In {\em CVPR}, 2020.

\bibitem{chan2020understanding}
Kelvin~CK Chan, Xintao Wang, Ke Yu, Chao Dong, and Chen~Change Loy.
\newblock Understanding deformable alignment in video super-resolution.
\newblock In {\em AAAI}, 2021.

\bibitem{chan2021basicvsr++}
Kelvin~CK Chan, Shangchen Zhou, Xiangyu Xu, and Chen~Change Loy.
\newblock Basicvsr++: Improving video super-resolution with enhanced
  propagation and alignment.
\newblock In {\em CVPRW}, 2021.

\bibitem{charbonnier1994two}
Pierre Charbonnier, Laure Blanc-Feraud, Gilles Aubert, and Michel Barlaud.
\newblock Two deterministic half-quadratic regularization algorithms for
  computed imaging.
\newblock In {\em CVPR}, 1994.

\bibitem{dong2014learning}
Chao Dong, Chen~Change Loy, Kaiming He, and Xiaoou Tang.
\newblock Learning a deep convolutional network for image super-resolution.
\newblock In {\em ECCV}, 2014.

\bibitem{fuoli2022fast}
Dario Fuoli, Martin Danelljan, Radu Timofte, and Luc Van~Gool.
\newblock Fast online video super-resolution with deformable attention pyramid.
\newblock {\em CoRR}, abs/2202.01731, 2022.

\bibitem{Fuoli-arxiv19-rlsp}
Dario Fuoli, Shuhang Gu, and Radu Timofte.
\newblock Efficient video super-resolution through recurrent latent space
  propagation.
\newblock {\em CoRR}, abs/1909.08080, 2019.

\bibitem{haris2018deep}
Muhammad Haris, Gregory Shakhnarovich, and Norimichi Ukita.
\newblock Deep back-projection networks for super-resolution.
\newblock In {\em CVPR}, 2018.

\bibitem{haris2019recurrent}
Muhammad Haris, Gregory Shakhnarovich, and Norimichi Ukita.
\newblock Recurrent back-projection network for video super-resolution.
\newblock In {\em CVPR}, 2019.

\bibitem{huang2015bidirectional}
Yan Huang, Wei Wang, and Liang Wang.
\newblock Bidirectional recurrent convolutional networks for multi-frame
  super-resolution.
\newblock In {\em NeurIPS}, 2015.

\bibitem{isobe2020video-eccv}
Takashi Isobe, Xu Jia, Shuhang Gu, Songjiang Li, Shengjin Wang, and Qi Tian.
\newblock Video super-resolution with recurrent structure-detail network.
\newblock In {\em ECCV}, 2020.

\bibitem{isobe2020video}
Takashi Isobe, Songjiang Li, Xu Jia, Shanxin Yuan, Gregory Slabaugh, Chunjing
  Xu, Ya-Li Li, Shengjin Wang, and Qi Tian.
\newblock Video super-resolution with temporal group attention.
\newblock In {\em CVPR}, 2020.

\bibitem{isobe2020revisiting}
Takashi Isobe, Fang Zhu, Xu Jia, and Shengjin Wang.
\newblock Revisiting temporal modeling for video super-resolution.
\newblock In {\em BMVC}, 2020.

\bibitem{jing2021turning}
Yongcheng Jing, Yiding Yang, Xinchao Wang, Mingli Song, and Dacheng Tao.
\newblock Turning frequency to resolution: Video super-resolution via event
  cameras.
\newblock In {\em CVPR}, 2021.

\bibitem{jo2018deep}
Younghyun Jo, Seoung Wug~Oh, Jaeyeon Kang, and Seon Joo~Kim.
\newblock Deep video super-resolution network using dynamic upsampling filters
  without explicit motion compensation.
\newblock In {\em CVPR}, 2018.

\bibitem{kim2016accurate}
Jiwon Kim, Jung~Kwon Lee, and Kyoung~Mu Lee.
\newblock Accurate image super-resolution using very deep convolutional
  networks.
\newblock In {\em CVPR}, 2016.

\bibitem{kim2016deeply}
Jiwon Kim, Jung~Kwon Lee, and Kyoung~Mu Lee.
\newblock Deeply-recursive convolutional network for image super-resolution.
\newblock In {\em CVPR}, 2016.

\bibitem{kingma2014adam}
Diederik~P. Kingma and Jimmy Ba.
\newblock Adam: {A} method for stochastic optimization.
\newblock In {\em ICLR}, 2015.

\bibitem{Lai-cvpr17-LapSRN}
Wei-Sheng Lai, Jia-Bin Huang, Narendra Ahuja, and Ming-Hsuan Yang.
\newblock Deep laplacian pyramid networks for fast and accurate
  super-resolution.
\newblock In {\em CVPR}, 2017.

\bibitem{lee2021dynavsr}
Suyoung Lee, Myungsub Choi, and Kyoung~Mu Lee.
\newblock Dynavsr: Dynamic adaptive blind video super-resolution.
\newblock In {\em WACV}, 2021.

\bibitem{li2019fast}
Sheng Li, Fengxiang He, Bo Du, Lefei Zhang, Yonghao Xu, and Dacheng Tao.
\newblock Fast spatio-temporal residual network for video super-resolution.
\newblock In {\em CVPR}, 2019.

\bibitem{li2020mucan}
Wenbo Li, Xin Tao, Taian Guo, Lu Qi, Jiangbo Lu, and Jiaya Jia.
\newblock Mucan: Multi-correspondence aggregation network for video
  super-resolution.
\newblock In {\em ECCV}, 2020.

\bibitem{li2021comisr}
Yinxiao Li, Pengchong Jin, Feng Yang, Ce Liu, Ming-Hsuan Yang, and Peyman
  Milanfar.
\newblock Comisr: Compression-informed video super-resolution.
\newblock In {\em ICCV}, 2021.

\bibitem{lin2021fdan}
Jiayi Lin, Yan Huang, and Liang Wang.
\newblock Fdan: Flow-guided deformable alignment network for video
  super-resolution.
\newblock {\em CoRR}, abs/2105.05640, 2021.

\bibitem{liu2013bayesian}
Ce Liu and Deqing Sun.
\newblock On bayesian adaptive video super resolution.
\newblock {\em IEEE transactions on pattern analysis and machine intelligence},
  36(2):346--360, 2013.

\bibitem{liu2017robust}
Ding Liu, Zhaowen Wang, Yuchen Fan, Xianming Liu, Zhangyang Wang, Shiyu Chang,
  and Thomas Huang.
\newblock Robust video super-resolution with learned temporal dynamics.
\newblock In {\em ICCV}, 2017.

\bibitem{liu2021large}
Hongying Liu, Peng Zhao, Zhubo Ruan, Fanhua Shang, and Yuanyuan Liu.
\newblock Large motion video super-resolution with dual subnet and multi-stage
  communicated upsampling.
\newblock In {\em AAAI}, 2021.

\bibitem{loshchilov2016sgdr}
Ilya Loshchilov and Frank Hutter.
\newblock Sgdr: Stochastic gradient descent with warm restarts.
\newblock {\em CoRR}, abs/1608.03983, 2016.

\bibitem{mu2020graph}
Jing Mu, Ruiqin Xiong, Xiaopeng Fan, Dong Liu, Feng Wu, and Wen Gao.
\newblock Graph-based non-convex low-rank regularization for image compression
  artifact reduction.
\newblock {\em IEEE Transactions on Image Processing}, 29:5374--5385, 2020.

\bibitem{nah2019ntire}
Seungjun Nah, Sungyong Baik, Seokil Hong, Gyeongsik Moon, Sanghyun Son, Radu
  Timofte, and Kyoung Mu~Lee.
\newblock Ntire 2019 challenge on video deblurring and super-resolution:
  Dataset and study.
\newblock In {\em CVPRW}, 2019.

\bibitem{ranjan2017optical}
Anurag Ranjan and Michael~J Black.
\newblock Optical flow estimation using a spatial pyramid network.
\newblock In {\em CVPR}, 2017.

\bibitem{sajjadi2018frame}
Mehdi~SM Sajjadi, Raviteja Vemulapalli, and Matthew Brown.
\newblock Frame-recurrent video super-resolution.
\newblock In {\em CVPR}, 2018.

\bibitem{tao2017detail}
Xin Tao, Hongyun Gao, Renjie Liao, Jue Wang, and Jiaya Jia.
\newblock Detail-revealing deep video super-resolution.
\newblock In {\em ICCV}, 2017.

\bibitem{tian2020tdan}
Yapeng Tian, Yulun Zhang, Yun Fu, and Chenliang Xu.
\newblock Tdan: Temporally-deformable alignment network for video
  super-resolution.
\newblock In {\em CVPR}, 2020.

\bibitem{wang2019deformable}
Hua Wang, Dewei Su, Chuangchuang Liu, Longcun Jin, Xianfang Sun, and Xinyi
  Peng.
\newblock Deformable non-local network for video super-resolution.
\newblock {\em IEEE Access}, pages 177734--177744, 2019.

\bibitem{wang2021tdn}
Limin Wang, Zhan Tong, Bin Ji, and Gangshan Wu.
\newblock Tdn: Temporal difference networks for efficient action recognition.
\newblock In {\em CVPR}, 2021.

\bibitem{wang2016temporal}
Limin Wang, Yuanjun Xiong, Zhe Wang, Yu Qiao, Dahua Lin, Xiaoou Tang, and Luc
  Van~Gool.
\newblock Temporal segment networks: Towards good practices for deep action
  recognition.
\newblock In {\em ECCV}, 2016.

\bibitem{wang2019edvr}
Xintao Wang, Kelvin~CK Chan, Ke Yu, Chao Dong, and Chen Change~Loy.
\newblock Edvr: Video restoration with enhanced deformable convolutional
  networks.
\newblock In {\em CVPRW}, 2019.

\bibitem{wei2021unsupervised}
Yunxuan Wei, Shuhang Gu, Yawei Li, Radu Timofte, Longcun Jin, and Hengjie Song.
\newblock Unsupervised real-world image super resolution via domain-distance
  aware training.
\newblock In {\em CVPR}, 2021.

\bibitem{xue2019video}
Tianfan Xue, Baian Chen, Jiajun Wu, Donglai Wei, and William~T Freeman.
\newblock Video enhancement with task-oriented flow.
\newblock {\em International Journal of Computer Vision}, 127(8):1106--1125,
  2019.

\bibitem{yi2021omniscient}
Peng Yi, Zhongyuan Wang, Kui Jiang, Junjun Jiang, Tao Lu, Xin Tian, and Jiayi
  Ma.
\newblock Omniscient video super-resolution.
\newblock In {\em ICCV}, 2021.

\bibitem{yi2019progressive}
Peng Yi, Zhongyuan Wang, Kui Jiang, Junjun Jiang, and Jiayi Ma.
\newblock Progressive fusion video super-resolution network via exploiting
  non-local spatio-temporal correlations.
\newblock In {\em ICCV}, 2019.

\bibitem{ying2020deformable}
Xinyi Ying, Longguang Wang, Yingqian Wang, Weidong Sheng, Wei An, and Yulan
  Guo.
\newblock Deformable 3d convolution for video super-resolution.
\newblock {\em IEEE Signal Processing yi2021omniscientLetters}, 27:1500--1504,
  2020.

\bibitem{zhang2018image}
Yulun Zhang, Kunpeng Li, Kai Li, Lichen Wang, Bineng Zhong, and Yun Fu.
\newblock Image super-resolution using very deep residual channel attention
  networks.
\newblock In {\em ECCV}, 2018.

\end{thebibliography}
}

\end{document}